\documentclass[sigconf, nonacm]{acmart}

\newcommand\vldbdoi{XX.XX/XXX.XX}
\newcommand\vldbpages{XXX-XXX}
\newcommand\vldbvolume{18}
\newcommand\vldbissue{1}
\newcommand\vldbyear{2024}
\newcommand\vldbauthors{\authors}
\newcommand\vldbtitle{\shorttitle} 
\newcommand\vldbavailabilityurl{https://github.com/Ryanhilde/dim_sum_vldb}
\newcommand\vldbpagestyle{plain}

\usepackage{amsmath,amssymb,amsfonts}\usepackage{graphicx}
\usepackage{textcomp}
\usepackage{xcolor}
\usepackage{caption}
\usepackage{subcaption}
\usepackage{tabularx}
\usepackage{booktabs}
\usepackage{easy-todo}
\usepackage{titlesec}
\usepackage{enumitem}
\usepackage{fancyhdr}
\usepackage{svg}
\usepackage{tikz}
\usepackage{booktabs}
\usepackage{adjustbox}
\usepackage{amsmath}
\usepackage{hyperref}

\usetikzlibrary{shapes,shapes.geometric,arrows,arrows.meta,positioning,calc,fit,backgrounds,decorations.pathreplacing}
\usepackage{tcolorbox}
\usepackage{pgfplots}
\usepackage{pgfplotstable}
\usepgfplotslibrary{groupplots}
\usepackage{multirow} 
\pgfplotsset{compat=1.17}
\usepackage{algpseudocode}
\usepackage{changepage}
\usepackage[linesnumbered,ruled,vlined]{algorithm2e}

\usepackage{xifthen} 

\usepackage{macros}

\begin{document}

\title{DIM-SUM: Dynamic IMputation for Smart Utility Management}

\author{Ryan Hildebrant}\affiliation{\institution{University of California, Irvine}}\email{rhildebr@uci.edu}\author{Rahul Bhope}\affiliation{\institution{University of California, Irvine}}\email{rbhope@uci.edu}\author{Sharad Mehrotra}\affiliation{\institution{University of California, Irvine}}\email{sharad@ics.uci.edu}\author{Christopher Tull}\affiliation{\institution{California Data Collaborative}}\email{chris@thecadc.org}\author{Nalini Venkatasubramanian}\affiliation{\institution{University of California, Irvine}}\email{nalini@uci.edu}

\begin{abstract}

Time series imputation models have traditionally been developed using complete datasets with artificial masking patterns to simulate missing values. However, in real-world infrastructure monitoring, practitioners often encounter datasets where large amounts of data are missing and follow complex, heterogeneous patterns. We introduce DIM-SUM, a preprocessing framework for training robust imputation models that bridges the gap between artificially masked training data and real missing patterns. DIM-SUM combines pattern clustering and adaptive masking strategies with theoretical learning guarantees to handle diverse missing patterns actually observed in the data. Through extensive experiments on over 2 billion readings from California water districts, electricity datasets, and benchmarks, we demonstrate that DIM-SUM outperforms traditional methods by reaching similar accuracy with lower processing time and significantly less training data. When compared against a large pre-trained model, DIM-SUM averages 2x higher accuracy with significantly less inference time.


\end{abstract}

\maketitle

\pagestyle{\vldbpagestyle}
\begingroup\small\noindent\raggedright\textbf{PVLDB Reference Format:}\\
\vldbauthors. \vldbtitle. PVLDB, \vldbvolume(\vldbissue): \vldbpages, \vldbyear.\\
\href{https://doi.org/\vldbdoi}{doi:\vldbdoi}
\endgroup
\begingroup
\renewcommand\thefootnote{}\footnote{\noindent
This work is licensed under the Creative Commons BY-NC-ND 4.0 International License. Visit \url{https://creativecommons.org/licenses/by-nc-nd/4.0/} to view a copy of this license. For any use beyond those covered by this license, obtain permission by emailing \href{mailto:info@vldb.org}{info@vldb.org}. Copyright is held by the owner/author(s). Publication rights licensed to the VLDB Endowment. \\
\raggedright Proceedings of the VLDB Endowment, Vol. \vldbvolume, No. \vldbissue\ %
ISSN 2150-8097. \\
\href{https://doi.org/\vldbdoi}{doi:\vldbdoi} \\
}\addtocounter{footnote}{-1}\endgroup

\ifdefempty{\vldbavailabilityurl}{}{
\vspace{.3cm}
\begingroup\small\noindent\raggedright\textbf{PVLDB Artifact Availability:}\\
The source code, data, and/or other artifacts have been made available at \url{\vldbavailabilityurl}.
\endgroup
}

\section{Introduction}
 
This paper considers the challenge of learning imputation models for large univariate time series datasets in the wild, where the input dataset may contain a significant number of missing values. Our motivation stems from the domain of civil infrastructure monitoring (e.g. water utilities, energy metering, intelligent transportation) where time series data is collected by multiple organizations over extended periods at fixed intervals. Such datasets frequently contain substantial missing values from factors such as sensor failures, communication interruptions, system maintenance, and software issues. For example, in our analysis of water data in California, we observed a water district with 47\% of the expected values missing. 

Large-scale missing data issues arise across various infrastructure datasets. Electric grid monitoring systems frequently experience significant gaps in data collection, particularly during periods of high load when data is most valuable. Studies by \cite{9016055, kuppannagari2021spatio} found that 47.14\% of individual readings can be missing, with 51.26\% of readings  being up to a half a day of continuous missing values. Similarly, loop sensor from the CA Dept. of Transportation's PEMS system \cite{chen2002quality} exhibits wide variations in missing data, with reporting areas experiencing between 38\% to 76\% missing values.

Beyond the sheer volume of missing data, infrastructure datasets often exhibit distinct missing patterns that correlate with specific operational factors. For instance, water meters from the same manufacturer may exhibit synchronized data gaps due to shared maintenance schedules or minimum required signal strengths for data transmission. In the electric domain, \cite{kuppannagari2021spatio} reported that understanding the relationship between voltage fluctuations and downstream load consumption in power grids improves the characterization and imputation of missing data. Given the level of missing values and the domain knowledge required, applying plug-and-play imputation techniques across infrastructure domains is challenging.





Traditional approaches to learning imputation models begin by dividing time series data into fixed-length windows (e.g., hourly periods) to capture temporal patterns and dependencies. A window is considered complete if it contains all expected measurements within its interval \cite{grail2021}. For example, if sensors report readings every 15 seconds, a complete 1-minute window would contain four measurements at their expected timestamp intervals. These complete windows serve as training data, where values are artificially masked (removed), and models are trained to reconstruct these masked values. However, in the wild it can be challenging to find a sufficiently large clean dataset with complete windows for training.

A comprehensive survey of time series imputation models, including various machine learning and deep learning approaches, can be found in \cite{tkde2022survey, thakur2021survey}. Recent machine learning advances have demonstrated significant improvements through several architectures: SAITS~\cite{du2023saits} employs self-attention mechanisms for robust imputation; non-stationary transformers~\cite{liu2022non} are designed to handle varying temporal dependencies; MRNN~\cite{yoon2018estimating} implements multi-directional recurrent neural networks for complex temporal relationships in streaming data with missing values; StemGNN~\cite{cao2020stemgnn} utilizes a graph neural network with a spectral GNN for capturing inter-series correlations and temporal dependencies; and CSDI~\cite{tashiro2021csdi} employs a conditional score-based diffusion model for imputing missing values by learning to reverse a noising process.

While these sophisticated architectures offer valuable insights into handling missing data, they often treat missing values as random masks or assume patterns drawn from predefined distributions, rather than as complex, domain-influenced structures. While this strategy has been effective in domains such as air quality, healthcare, and activity recognition~\cite{yoon2018estimating}, it presents challenges in large-scale infrastructure settings, where there is large datasets with significant amounts of missing values. These settings are characterized by numerous missing patterns that, despite their complexity, are recorded and reported at fixed intervals. Thus, an underlying assumption in our work is that we operate with a complete dataset. This means we presuppose a fixed data interval and an expected quantity of data points within that interval.

Several innovations have explicitly addressed how to represent large and diverse patterns of missing data to feed into an imputation model. A significant advancement is DAGAN~\cite{liu2021adaptive}, which utilizes two Generative Adversarial Networks (GANs) to learn and replicate missing patterns from real-world datasets. By training on actual production data rather than synthetic patterns, DAGAN aims to model the complexity of real-world missingness. It introduces a method of "projecting" test set values with missing patterns onto the training set to create pattern-specific masks for model training. Another strategy involves consolidating multiple datasets into a single, large pretrained model capable of processing substantial data and adapting to various missing data patterns, such as Amazon's Chronos \cite{ansari2024chronos}. Both of these techniques fundamentally assume that the entire dataset is known upfront. This includes both observable data and what might be termed "non-observable patterns"—essentially, the awareness that an expected value is missing, thereby reducing the anticipated amount of data. Methodologies designed to leverage real patterns of missing data inherently rely on this comprehensive, upfront knowledge of the dataset.

Hence, our goal is to explore how imputation models can be aware of real missing patterns, and "bake" them into the training process. Previous work has focused on explicitly defined patterns, single sources of missing data, or are assumed to be completely known during training time (e.g., from production datasets~\cite{liu2021adaptive}). In contrast, infrastructure time series settings often have implicit missing patterns that can be numerous and often cluster based on characteristics such as meter type and land-use, e.g., single-family homes exhibit different usage and missing patterns than industrial factories or multi-family homes.
 
To address this challenge, we develop \textbf{DIM-SUM}: a generic framework for baking missing patterns into time series imputation models. DIM-SUM introduces a model-agnostic framework that enhances existing imputation methods by systematically incorporating missing patterns discovered through data analysis. Rather than developing new architectures or merging existing approaches, DIM-SUM provides a methodology that allows any imputation model to learn from and adapt to varying patterns of missing data. This enables practitioners to select models based on their specific data needs while maintaining robust performance across different missing scenarios and significantly reducing the amount of data that needs to be used for training. \textbf{Our contributions are as follows:}
\begin{itemize}
    \item \textbf{We propose a technique to preprocess imputation models for large time series datasets using limited training data}. DIM-SUM provides a framework that can be applied with any existing imputation model (Section \ref{sec:framework}).
    
    \item \textbf{We introduce a methodology for baking data from clean and dirty sources} and projecting them into training data with minimal additional noise (Sections~\ref{sec:clustering} \& \ref{sec:training}).
    
    \item \textbf{We provide probabilistic bounds for the quality of trained imputation models given diverse data sources} and unknown patterns of missing data (Section~\ref{sec:pac_analysis}).
    
    \item \textbf{We evaluate our framework on six large-scale  datasets from water, electric, and weather domains}, comparing DIM-SUM against several baselines: (i) a direct model approach using fixed missing patterns, (ii) GAN-based methods that simulate various missing patterns, and (iii) pre-trained large language models (Section~\ref{sec:eval}).
\end{itemize}
\begin{figure*}
    \includegraphics[width=0.99\textwidth]{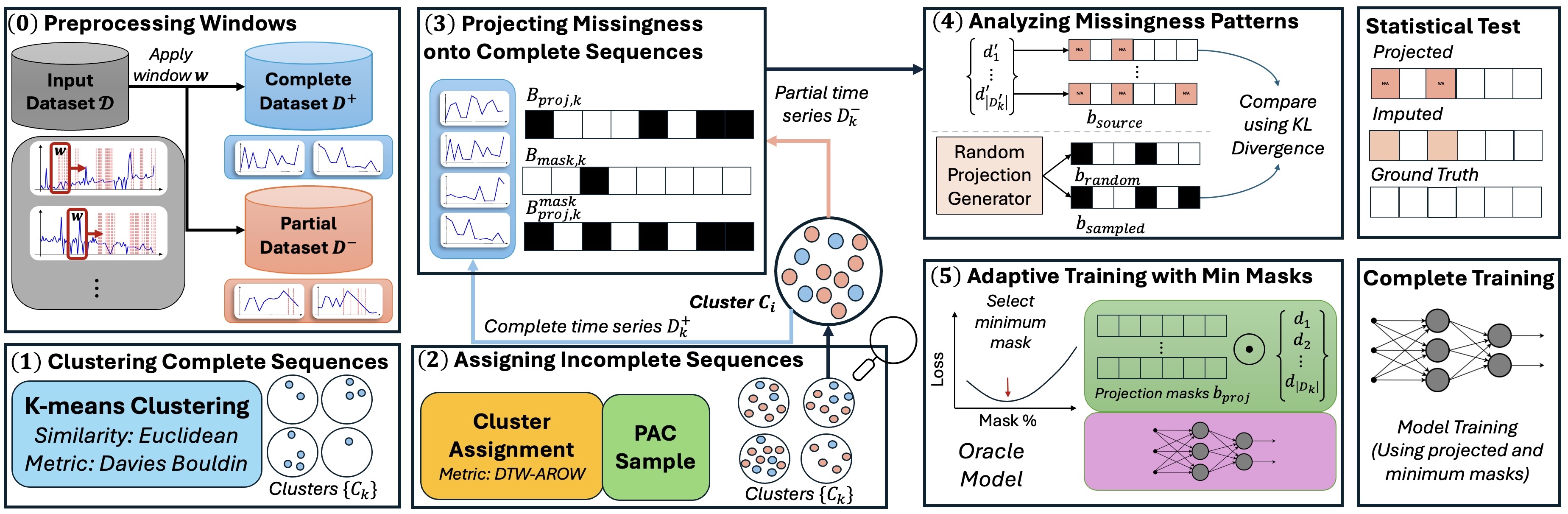}
    \caption{Overview of the DIM-SUM Preprocessing (0, 1, 2) and Training (3, 4, 5) Framework for Time Series Imputation}
    \label{fig:dim_sum}
    \Description{}
\end{figure*}

\section{Related Work}
\label{sec:related_work}

The study of missing data patterns is well-established~\cite{Little2002,de2011handbook,khayati2020mind,miao2022efficient}, with statistics categorizing missing data as Missing Completely at Random (MCAR), Missing at Random (MAR), and Missing Not at Random (MNAR) to describe the relationship between missingness and data values~\cite{LittleRubin2002, Rubin1976}. In infrastructure time series, instances of missing values are typically known due to the data's periodic nature~\cite{Rubin1976}. While traditional imputation methods often assume MCAR (where missingness is independent of data values), infrastructure settings frequently exhibit complex MNAR patterns (e.g., sensor failures due to extreme conditions, where missingness relates to unobserved values) or MAR patterns (e.g., scheduled maintenance, where missingness depends on observed variables).

Addressing these challenges, traditional database systems have adapted and scaled statistical methods for large-scale time series processing~\cite{khayati2024imputevis, sigmod2021survey, vldb2021survey, tkde2022survey,khayati2021orbits, lubba2019catch22}. Early systems integrated imputation directly into query processing, such as ImputeDB~\cite{cambronero2017query}, while later frameworks like HoloClean~\cite{rekatsinas2017holoclean} introduced scalable probabilistic data repair. These evolved into distributed systems like DAME~\cite{dame2020} and EDIT~\cite{miao2022efficient}, which implement variants of classical imputation methods while maintaining statistical guarantees~\cite{lin2023zip, fan2014towards}. Furthermore, approaches like GRAIL~\cite{grail2021} have focused on efficient time-series representation learning and processing at scale, informing scalable imputation strategies.

Deep learning has significantly advanced imputation by capturing complex temporal dependencies~\cite{wu2023timesnet}. Recurrent Neural Networks (RNNs) like BRITS~\cite{cao2018brits}, GRUD~\cite{che2018recurrent}, and MRNN~\cite{yoon2018estimating} utilized bidirectional structures and mechanisms for variable-length or streaming data. Transformer architectures, including SAITS~\cite{du2023saits}, non-stationary transformers~\cite{liu2022non}, ETSformer~\cite{Voorhees2022ETSformer}, and DAMR~\cite{ren2023damr}, have effectively modeled long dependencies using self-attention~\cite{zhang2023crossformer}. Other innovations include ImDiffusion~\cite{chen2023imdiffusion} using diffusion models, CSDI~\cite{tashiro2021csdi} employing conditional score-based diffusion, Series2Graph~\cite{boniol2020series2graph} combining imputation with graph representations, StemGNN~\cite{cao2020stemgnn} leveraging spectral graph neural networks, uncertainty-driven networks~\cite{feng2024missing} aiming for efficiency, and approaches like DeepMVI~\cite{bansal2021missing} focusing on multidimensional time series.

DAGAN~\cite{liu2021adaptive} made strides in incorporating realistic missing patterns but has limitations for infrastructure data. It learns mappings from a single clean dataset to multiple missing patterns, necessitating separate models for each source and struggling with the scale of millions or billions of readings. This approach is computationally challenging for infrastructure settings that require handling multiple source patterns mapping to multiple target patterns within a unified, efficient framework. Furthermore, DAGAN's GAN-based architecture lacks theoretical guarantees regarding the fidelity of learned missing patterns or their reproduction during training.

\section{The DIM-SUM Approach and Architecture}
\label{sec:framework}

Figure~\ref{fig:dim_sum} illustrates the various steps of DIM-SUM and how the components work together to create a robust imputation framework. Algorithm \ref{alg:preprocessing} describes the data preparation and clustering steps (0, 1 and 2) in DIM-SUM, while Algorithm \ref{alg:training} specifies the model training steps (3, 4, and 5) in the DIM-SUM learning technique. A more detailed overview of \ref{alg:preprocessing} can be found in Section \ref{sec:clustering} and a discussion of the training process is found in \ref{sec:training}.

\textbf{0. Windowing Function:} DIM-SUM begins by ingesting a collection of univariate time series $\dataset{}{} = \{\TS{1}{}, \ldots, \TS{N}{} \}$, where each sequence $\TS{i}{} \in \mathbb{R}^{\nTimestep}$ consists of $\nTimestep$ time steps that are partitioned into smaller, fixed-length windows (Algorithm \ref{alg:preprocessing}, lines 2-5). This step standardizes sequence lengths, accommodates varying sampling rates, and ensures that imputation models operate on locally consistent patterns. Specifically, we apply a tumbling window function of size $\window$, resulting in each sequence $\TS{i}{}$ being divided into $\frac{\nTimestep}{\window}$ non-overlapping segments $\TS{i}{j}$. If a sequence contains 1,000 readings and $\window = 100$, it is split into 10 partitions. Each window retains an associated mask $\mask{i}{j} \in \{0,1\}^{\window}$, where $1$ indicates an observed value and $0$ denotes a missing value. After windowing, we construct two subsets: $\completeTScluster{} $, containing fully observed windows, and $\incompleteTScluster{}$, containing windows with missing values, where $|\completeTScluster{}| \ll |\incompleteTScluster{}|$.

\textbf{1. Clustering Complete Sequences.} To introduce structure in the training data, we first cluster fully observed sequences in $\completeTScluster{}$ (Algorithm \ref{alg:preprocessing}, lines 6-7). Since time series data can exhibit highly variable patterns, direct imputation without accounting for inherent structure may yield poor results. We apply mini-batch $\nCluster$-means clustering to identify dominant patterns and select the number of clusters $\nCluster$ using the Davies-Bouldin index~\cite{davies1979cluster}, which is determined by passing $D^+$ to $DBIFindOptimalK$. The clustering process aims to group similar temporal patterns, providing a structure for imputing missing values in $\incompleteTScluster{}$.

\textbf{2. Incomplete Cluster Assignment.} Once the complete sequences have been clustered, we assign the incomplete sequences in $\incompleteTScluster{}$ to the closest cluster. A naive approach would involve simple distance-based matching, but this can be unreliable when dealing with sequences that contain missing values. Instead, we employ DTW-AROW $\DTWAROW{\TS{i}{j}}{\TS{p}{q}}$, a dynamic time warping (DTW)-based method that preserves temporal alignment while handling missing values effectively~\cite{yurtman2023estimating, sart2010accelerating}. For each incomplete sequence in $D^-$, we find the closest fit to the existing complete cluster centroids (Algorithm \ref{alg:preprocessing}, lines 8-9). This approach ensures that incomplete sequences are mapped to clusters that best reflect their structure. To improve assignment stability, we apply a PAC-bound based sampling heuristic~\cite{vapnik1998statistical}, which prevents clusters from becoming unbalanced by distributing sequences more uniformly, which we describe in detail in Section \ref{sec:min_mask}.

\SetKw{break}{break}
\begin{algorithm}[!ht]
\small
\KwIn{Dataset $\dataset{}{}$, window size $\window$, cluster range $(\nCluster_{\text{min}}, \nCluster_{\text{max}})$, PAC bound $\tau$}
\KwOut{Clusters $\{\TScluster{k}\}$ with assigned sequences}
$\completeTScluster{} \gets \emptyset$, $\incompleteTScluster{} \gets \emptyset$; \\
\tcp{Step 0: Window partitioning}
\For{$\TS{i}{} \in \dataset{}{}$}{
    $\{\TS{i}{1}, \ldots, \TS{i}{\lfloor \nTimestep / \window \rfloor}\} \gets \text{WindowPartition}(\TS{i}{}, \window)$\\
    \For{each window $\TS{i}{j}$}{
        \leIf{$\text{IsComplete}(\TS{i}{j})$}{
            $\completeTScluster{} \gets \completeTScluster{} \cup \{\TS{i}{j}\}$
        }{
            $\incompleteTScluster{} \gets \incompleteTScluster{} \cup \{\TS{i}{j}\}$
        }
    }
}
\tcp{Step 1: Complete sequence clustering}
$\nCluster \gets \text{DBIFindOptimalK}(\completeTScluster{}, \nCluster_{\text{min}}, \nCluster_{\text{max}})$\\
$\{\TScluster{k}\} \gets \text{ClusterSequences}(\completeTScluster{}, \nCluster)$

\tcp{Step 2: Assign incomplete sequences}

\For{$D^-_k \in \incompleteTScluster{}$}{
    $D^-_k \gets IncompleteClusterAssignment(D^- , C_k, \tau)$}

\Return{$\{\TScluster{k}\}$}
\caption{DIM-SUM Preprocessing \& Clustering}
\label{alg:preprocessing}
\end{algorithm}

\begin{algorithm}[!ht]
\small
\KwIn{Clusters $\{\TScluster{k}\}$, assignments $\{\incompleteTScluster{k}\}$, model $\imputationmodel{}$}
\KwOut{Trained models $\{\imputationmodel{k}\}$}
\For{$k \gets 1 ~\KwTo~ \nCluster$}{
    \tcp{Step 3: Create projections}
    
    \For{$\TS{i}{j} \in \incompleteTScluster{k}$}{
        $\mask{i}{j} \gets \text{GetMask}(\TS{i}{j})$;
        
        $\TS{p}{q} \gets \text{SampleFrom}(\completeTScluster{k})$;
        
        $\maskproj \gets \ProjectionOperator{\TS{i}{j}, \TS{p}{q}}$;
        
        $\projectedTScluster[proj]{k} \gets \projectedTScluster[proj]{k} \cup \{\maskproj\}$
    }
    \tcp{Step 4: Analyze structure}
    $n_p \gets 0$\;
    \For{$\masksrc \in \projectedTScluster[proj]{k}$}{
        $\masksamp \gets \text{SamplePattern}(\incompleteTScluster{k})$\;
        $\maskrand \gets \text{RandomPattern}()$\;
        \lIf{$\KLdivergence{\masksrc \parallel \masksamp} < \KLdivergence{\masksrc \parallel \maskrand}$}{
            $n_p \gets n_p + 1$
        }
    }
    
    \tcp{Step 5: Adaptive training}
    $MinMaskSearch(m_{min}, \alpha, \omega, T)$
}
\Return{$\{\imputationmodel{k}\}$}
\caption{DIM-SUM Training}
\label{alg:training}
\end{algorithm}

\textbf{3. Projecting Missing Patterns onto Complete Sequences.}
To ensure that imputation models are trained in conditions that reflect real-world missing patterns, we generate realistic incomplete sequences by projecting observed missing patterns onto fully observed data. For each cluster $k$, given an incomplete sequence $\TS{i}{j} \in \incompleteTScluster{k}$, we extract the missing patterns $\mask{i}{j} \in \{0,1\}^{\window}$. These binary sequences captures the distribution of missing values in the time series (Algorithm \ref{alg:training}, lines 1-2). We then apply this pattern to a randomly selected complete sequence within the same cluster $\TS{p}{q} \in \completeTScluster{k}$, effectively masking values where elements of $\mask{i}{j}$ is $0$. This process generates a new dataset of masked sequences $D_{p,k}$ that retain the structural properties of complete data while incorporating  missing values from that cluster's distribution. Each masked sequence $\masksrc \in \projectedTScluster[proj]{k}$ is created through the projection operation $\ProjectionOperator{}$ applied to the sequence pair (Algorithm \ref{alg:training}, lines 3-7).

\textbf{4. Analyzing missing patterns}
Once the projected datasets $\projectedTScluster[proj]{k}$ have been constructed for each cluster $\TScluster{k}$, we analyze whether the partial sequences have similarities between the centroids of each cluster. Within a cluster, we compare each observed missing pattern $\masksrc \in \projectedTScluster[proj]{k}$ against both a sampled pattern $\masksamp$ from another sequence in $\incompleteTScluster{k}$ and a randomly generated missing pattern $\maskrand$. If the two samples from the cluster exhibit a structured relationship with the underlying behavioral pattern, imputation should account for these dependencies rather than treating missing values as independent noise (Algorithm \ref{alg:training}, lines 9-12).

We measure the structured relationship as the similarity between observed and reference missing patterns using the KL divergence metric \cite{kullback1951information}: $\KLdivergence{\masksrc \parallel \masksamp}$ and $\KLdivergence{\masksrc{}{} \parallel \maskrand}$
If a majority of the missing patterns in $\projectedTScluster[proj]{k}$ demonstrate non-random structure, the cluster exhibits similar missing data behavior. This informs the subsequent training of cluster-specific models $\imputationmodel{k}$.

\textbf{5. Adaptive Training with Minimum Masks.}
Finally, DIM-SUM employs an iterative training process that determines the minimum masking necessary for robust imputation. For each cluster $k$, we begin with an initial mask size $\masksize{0}$ and progressively increase the fraction of missing values introduced during training. With our projected patterns $\masksrc \in \projectedTScluster[proj]{k}$ already capturing real distributions, we create two additional representations: $\projectedTScluster[mask]{k} = \completeTScluster{k} \odot \masksrc$ applying only artificial masks to complete data, and $\projectedTScluster[proj,mask]{k} = \completeTScluster{k} \odot \masksrc$ applying artificial masks to windows of data. At each iteration, we train two models: an oracle model using only $\projectedTScluster[mask]{k}$, and a projection model using $\projectedTScluster[proj,mask]{k}$. If the projection model's performance remains within two standard deviations of the oracle model's performance while handling progressively larger missing regions, training continues; otherwise, the process is halted.

The details of this minimum mask selection process are detailed in Algorithm \ref{alg:mask_search} in Section \ref{sec:min_mask}. Once a minimum mask is selected by sampling the training data (determined by the PAC bound, which we describe later in Section~\ref{sec:pac_analysis}), we train the imputation model across the cluster-specific data and apply inference. We validate the performance of the model on the projected data using a statistical test to measure how well a model reconstructs missing values.

\section{Baking Missing Patterns via Clustering}
\label{sec:clustering}

Through our empirical studies with various smart utility datasets, we have found that the large scale and intricate nature of this data frequently results in these state-of-the-art imputation models typically produce unpredictable results.
This is particularly evident in the large water dataset comprising of 1.5 billion readings, where approximately 30\% of values are missing. When applying the Non-stationary Transformer model \cite{liu2022non} directly to the clean portions of this dataset and injecting 10\% of MCAR missing values, we observed an MSE of 1.4876. However, when we apply DIM-SUM to same transformer-based architecture, the MSE dramatically improved to 0.7770 (47.8\% error reduction).

Our performance gain comes from the recognition that utility consumption data contains repetitive behavioral patterns. Rather than treating the dataset as a monolithic entity, partitioning it into clusters of similar behavior allows us to train specialized smaller models that implicitly learn patterns of missing sequences on complete values that additionally have similarity based on what we can observe. By identifying and grouping similar temporal behaviors, these pattern-specific models better capture the distribution characteristics of each cluster, leading to more accurate imputations even with substantially fewer computational resources than using one large model on the entire dataset.

We identify distinct temporal patterns within complete windows $\completeTScluster{}$, where each window $\TS{i}{j}$ represents $w$ consecutive measurements. This requires balancing pattern specificity with having enough training samples per cluster. Too many clusters can create sparse sample distribution, while too few fail to capture temporal pattern diversity. We directly apply mini-batch $\nCluster$-means clustering to all windows, which reduces computational cost through batch processing while preserving clustering quality on large-scale time series data~\cite{sculley2010web}. Each window is first normalized using z-score normalization to focus on pattern shapes rather than absolute values.

Normalization ensures that clustering captures similarities in the temporal dynamics (the shape of the time series) rather than being dominated by magnitude differences. This helps identify meaningful patterns across sensors or devices with different baseline readings but similar behavioral patterns, regardless of scale. Such an approach is quite effective for the cyclic nature of utility usage, i.e., single family homes may take showers in the morning, but amount of water consumed may differ. To determine the optimal number of clusters, we perform a binary search over the range $[\nCluster_{\text{min}}, \nCluster_{\text{max}}]$, using the Davies-Bouldin index \cite{davies1979cluster} as our optimization criterion, lines 6-7 in Algorithm \ref{alg:preprocessing}. For each candidate $k$, we compute:
\begin{equation}
\text{DB}(k) = \frac{1}{k} \sum_{i=1}^k \max_{j \neq i} \left(\frac{\sigma_i + \sigma_j}{d_{ij}}\right)
\end{equation}

where $\sigma_i$ represents the average distance between points in cluster $i$ and their centroid, and $d_{ij}$ is the distance between the centroids of clusters $i$ and $j$. The Davies-Bouldin index measures the average similarity between each cluster and its most similar cluster, where similarity is defined as the ratio of within-cluster distances to between-cluster distances. Lower values indicate better clustering, with more compact and well-separated clusters.

We employ a binary search rather than an exhaustive search over all possible $k$ values to efficiently find the optimal number of clusters. This approach reduces the computational burden while still providing a robust estimate of the optimal $k$. Once the optimal number of clusters is determined in our sample space, we obtain the final clustering by applying $\text{ClusterSequences}(\completeTScluster{}, \nCluster)$. The optimal $K$ search has a time complexity of $\mathcal{O}(\log(R_K) \cdot (t K |D^{+}| w + K^2 w))$, for search range $R_K$, $t$ K-means iterations, $K$ clusters, $|D^{+}|$ complete windows, and window size $w$.

\subsection{Assigning Incomplete Sequences}
Training imputation models presents a fundamental question when working with limited complete data: \textit{how do we balance the use of complete sequences and utilize the observable data in the incomplete sequences for training?} This challenge is amplified when the data is partitioned into clusters to capture distinct temporal patterns, as it further reduces the available training samples per model. To address this limitation while maintaining the benefits of cluster-specific models, we develop a sampling strategy that augments each cluster's training data with incomplete sequences. While we cannot directly validate imputation quality on real missing values, we can project observed missing patterns onto complete sequences where reconstruction accuracy can be measured. Consider a complete time series sequence and a projected version of that time series with missing values written over random indices:
\[
\begin{aligned}
\text{Complete:} & \quad [10, 8, 7, 5, 6, 8, 12, 15, 14, 12, 10, 9, 8, 7, 8, 10, 13, 15, 14] \\
\text{Projection:} & \quad [10, \textcolor{red}{m}, 7, \textcolor{red}{m}, 6, 8, \textcolor{red}{m}, \textcolor{red}{m}, 14, 12, \textcolor{red}{m}, 9, 8, 7, \textcolor{red}{m}, 10, 13, \textcolor{red}{m}, 14]
\end{aligned}
\]
The projected sequence is fed into an model, which will output its best estimate of the original complete window. Although this is a standard practice, the projection does not account for the shared patterns of the observable values between complete and incomplete sequences. These subsequences of values also provide important information for fitting the right sequence to an existing cluster by exploiting the distributional closeness of observable values. 

For assigning incomplete sequences to a cluster, we use Dynamic Time Warping with Aligned Resolutions of Warping (DTW-AROW~\cite{yurtman2023estimating}), which handles sequences with missing values while preserving temporal relationships with an extended cost function:
\[
\delta(x_i, x_j) =
\begin{cases}
0, & \text{if } x_i = \text{NaN} \text{ or } x_j = \text{NaN} \\
(x_i - x_j)^2, & \text{otherwise}
\end{cases}
\]

DTW-AROW enforces synchronized advancement through missing regions and applies a correction factor, which is calculated as the ratio of observable values to all values in both sequences. This prevents artificial alignments while ensuring fair comparisons between sequences with different proportions of missing data. The computational complexity of comparing each incomplete sequence with all centroids would be $O(|D^-| \cdot |D^+|/C)$, where $C$ is the number of sequences per centroid. Given that $|D^+| \ll |D^-|$, we leverage sampling theory to reduce this to $O(n \cdot k)$, where $n$ is the number of sampled incomplete sequences and $k$ is the number of clusters.

To determine the required number of incomplete sequences, we leverage PAC learning theory \cite{vapnik1998statistical}, which is developed formally in Section~\ref{sec:pac_analysis}. For each cluster, we compute an observable value ratio $\gamma = (1-\alpha)(1-\mathbb{M})$, where $\alpha$ is the ratio of missing values that are currently present and $\mathbb{M}$ is the masking ratio used during training, as seen in Algorithm \ref{alg:partial_cluster} (lines 8-9, Algorithm \ref{alg:preprocessing}).  

The observable value ratio tells us how many true values the model receives during training, which we form our bound on. The goal of the cluster assignment algorithm is to sample enough missing data into each cluster, such that there is a strong trend of similar patterns to project. If there is not a strong trend, it indicates to us that the patterns are unlikely to be significant across the entire dataset. Furthermore, by sampling enough observable data to meet a PAC bound and assuming our data is independent and identically distributed (I.I.D), we can approximately ensure the expected performance achieved during training on the cluster-specific model translate to the test data that fits the properties of each cluster.

\begin{algorithm}
\caption{Incomplete Sequence Cluster Assignment}
\label{alg:partial_cluster}
\KwIn{Dataset $\incompleteTScluster{}$, clusters $\{C_k\}$, PAC threshold $\tau$, MaxClusterSize, $\alpha$, $\mathbb{M}$}
\KwOut{Augmented cluster assignments}
\For{$\TS{i}{j} \in \incompleteTScluster{}$}{
    $\gamma_i \gets (1-\alpha)(1-\mathbb{M})$\;
    \ForEach{$C_k$}{
        $d_k \gets \DTWAROW{\TS{i}{j}}{C_k}$\;
    }
    $S_c \gets \text{SortByDistance}(d_k)$\;
    \For{$c \in S_c$}{
        \If{$\text{Observable}_c < \tau$ \textbf{and} $|C_c| < \text{MaxClusterSize}$}{
            $C_c \gets C_c \cup \{\TS{i}{j}\}$\;
            Update $\alpha_i$ based on in $\TS{i}{j}$\;
            $\gamma_i \gets (1-\alpha_i)(1-\mathbb{M})$\;
            $\text{Observable}_c \gets \text{Observable}_c + \gamma_i$\;
            Update $\tau$ if necessary\;
            \textbf{break}\;
        }
    }
    \If{$\TS{i}{j}$ not assigned}{
        $c^* \gets \arg\min_{c \in S_c} |C_c|$\;
        $C_{c^*} \gets C_{c^*} \cup \{\TS{i}{j}\}$\;
    }
}
\Return{$\{C_k\}$}
\end{algorithm}

The algorithm initially calculates the PAC bound assuming $\alpha$ incomplete data. Once we sample this initial calculated amount, we examine the actual projected missing patterns, recalculate the bound, and add more samples if needed. This dynamic adjustment allows us to determine the number of samples needed to satisfy our learning guarantees more accurately.

The heuristic approach to cluster assignment first attempts to place an incomplete sequence in the closest matching cluster based on DTW-AROW distance. However, if this cluster has reached its capacity, we assign the sequence to the next closest available cluster. This ensures a more even distribution of incomplete sequences across clusters while still prioritizing similarity. The motivation for assigning to the next closest cluster is that it allows us to quickly sample and fill clusters with strong correlations to meet the bound. This helps reduce the amount of training samples we need to iterate through and fills less frequent clusters faster. In later steps, we show that this heuristic approach works well for identifying strong clusters and eliminating clusters which have no dominant pattern trends. Next, we can now prepare the data for training an imputation model that is fitted to the distributions of that particular cluster.
\section{Training Models with Incomplete Data}
\label{sec:training}

The clustering of time series based on their characteristics and missing patterns raises a practical question: \textit{How can imputation models be effectively trained using a mix of complete sequences $(\completeTScluster{k} \in D^+)$ and sequences with missing values $(\incompleteTScluster{k} \in D^-)$ within each cluster?} An effective training strategy must address two objectives:
\begin{enumerate}
    \item Leveraging real-world missing patterns present in $\incompleteTScluster{k}$ is useful to represent the actual scenarios that models will encounter in production environments. These patterns contain information about how and when data becomes unavailable, but provide limited data to the model.
    \item Acquiring sufficient complete data from $\completeTScluster{k}$ is required for learning the underlying temporal dynamics and distributions that govern each cluster. Complete sequences provide the ground truth necessary for understanding the range of valid behaviors and relationships within the data, but may limited in representing all distributions across the dataset. 
\end{enumerate}
The DIM-SUM framework addresses these challenges by combining information from both $\completeTScluster{k}$ and $\incompleteTScluster{k}$. We translate the missing patterns from $\incompleteTScluster{k}$ into binary encodings and project them onto the complete sequences in $\completeTScluster{k}$. The projection of missing patterns effectively creates regions of missing values in the training data that reflect real-world scenarios, i.e., the data is ``projected'' away from the complete sequences before training. This simulates an environment where the model must train on sequences with missing values, forcing it to adapt to incomplete information. 

During training, the projected missing values are treated as genuinely missing, requiring the model to learn how to handle missing data in the observable regions to compute the loss. This is achieved using a masking process. However, excessively masking the remaining observable values might introduce unnecessary noise into the training process, potentially harming model performance. To address this, the framework focuses on identifying a minimal effective mask—the smallest amount of additional masking necessary to ensure the model performs effectively.

Mask size is determined by comparing against an oracle model trained only on complete sequences with the same small mask (meaning all the projected values are observable during training), representing an ideal training situation. When a model trained with both projected missing values and masked values achieves performance within a reasonable threshold of this oracle, it indicates an appropriate balance between learning from missing patterns and preserving sufficient training data. We assert this balance to be the difference in loss achieved by oracle model (which only imputes the masked values) and the projection model (imputes masked and projected values) is minimal across the various mask percentages.This process takes $\mathcal{O}((|D^{-}| + \sum_k N_{\text{train},k}) \cdot w)$ time, for $|D^{-}|$ incomplete windows, $N_{\text{train},k}$ training sequences of length $w$ per cluster.

\subsection{Pattern Projection and Masking} For any given cluster $\TScluster{k}$, we have both a complete subset $\completeTScluster{k} \subset \completeTScluster{}$ containing windows of full time series and a partial subset $\incompleteTScluster{k} \subset \incompleteTScluster{}$ containing series with missing values. Our goal is to learn from both sets while ensuring we maintain theoretical guarantees about model performance. To utilize both sequences, we use a process of projection, where the missing patterns from windows in $\incompleteTScluster{k}$ are applied to complete windows in $\completeTScluster{k}$ to create training data with a ground truth. Ultimately, we ``project'' these values away as input to the model for training, simulating a scenario where the model is forced to train on incomplete training data, but we are able to verify it's accuracy after training since we have the ground truth. 

We define a projection $\ProjectionOperator{}$ as an operation that maps a missing patterns from a sequence in $\incompleteTScluster{k}$ onto a sequence in $\completeTScluster{k}$. This operation creates masked values in the complete sequence that mirror the missing values in the partial sequence, expressed as $\ProjectionOperator{} : \incompleteTScluster{k} \times \completeTScluster{k} \rightarrow \{0,1\}^{\window} \times \mathbb{R}^{\window}$, where $\window$ is the window length. For each sequence $\TS{i}{j} \in \incompleteTScluster{k}$ and a complete sequence $\TS{p}{q} \in \completeTScluster{k}$, we encode its missing patterns as a binary vector $\masksrc$ which equals 1 if a value is present and 0 if it is missing. The projection then creates a binary mask and applies it to a complete sequence: $\ProjectionOperator{\TS{i}{j}, \TS{p}{q}}$.

To maintain the theoretical guarantees established during cluster assignment, we ensure a one-to-one mapping between partial and complete sequences. Each partial sequence $\TS{i}{j}$ is projected onto exactly one complete sequence $\TS{p}{q}$ chosen uniformly at random, creating our projected dataset $\projectedTScluster[proj]{k} = \Bigl\{\, \ProjectionOperator{\TS{i}{j}, \TS{p}{q}} : 
\TS{i}{j} \in \incompleteTScluster{k}, \TS{p}{q} \in \completeTScluster{k} \,\Bigr\}$. This bijectivity preserves the PAC-bounded properties of our original cluster assignments.

After establishing these initial projections, we analyze whether the missing patterns within each cluster exhibit meaningful structure. We maintain a count $n_{\text{pattern}}$ of windows whose missing patterns are more similar to other patterns in $\incompleteTScluster{k}$ than to randomly generated ones ($n_p$ in Algorithm \ref{alg:training}). For each source pattern $\masksrc$, we use Kullback-Leibler (KL) divergence \cite{kullback1951information} to compare it against both a randomly sampled pattern from the same cluster $\masksamp$ and a synthetic pattern with matching missing rate $\maskrand$. This comparison is formalized through the following decision criterion:
\begin{equation}
\maskproj =
\begin{cases}
\masksamp & \text{if } \KLdivergence{\masksrc | \masksamp} < \KLdivergence{\masksrc | \maskrand} \\
\maskrand & \text{otherwise}
\end{cases}
\end{equation}
A cluster is considered to exhibit meaningful structure when more than two-thirds ($\omega$ in Algorithm \ref{alg:mask_search}) of its patterns show stronger similarity to other real patterns than to random patterns. We measure this by computing a KL divergence criterion: 
{\small
\begin{equation}
\frac{1}{\left|\projectedTScluster{k}\right|}
\sum\limits_{\masksrc \in \projectedTScluster[proj]{k}} 
\sum\limits_{\masksamp \sim \projectedTScluster[proj]{k}}
\!\!\!\!
\mathbb{I}
\left(
    \KLdivergence{\masksrc \!\!\parallel\!\! \masksamp}
    <
    \KLdivergence{\masksrc \!\!\parallel\!\! \maskrand}
\right)
\end{equation}
}
where $\mathbb{I}$ is the indicator function. For each pattern that satisfies this criterion, we increment $n_{\text{pattern}}$, providing a running measure of the cluster's structural significance. If the ratio exceeds $\omega$, it suggests a dominant missing pattern within the cluster. If the ratio is below $\omega$, the missing patterns are likely rare or represented elsewhere, prompting the removal of that cluster. For clusters meeting this criterion, we proceed to create distinct training representations:
\begin{equation}
    \begin{aligned}
    \projectedTScluster[proj]{k} &= \completeTScluster{k} \odot \maskproj \\
    \projectedTScluster[mask]{k} &= \completeTScluster{k} \odot \maskmask \\
    \projectedTScluster[proj,mask]{k} &= \completeTScluster{k} \odot (\maskproj \land \maskmask)
    \end{aligned}
\end{equation}
where $\odot$ represents element-wise masking and $\land$ denotes the logical AND operation. Throughout this process, we retain the ground truth values for all masked positions to enable proper evaluation.

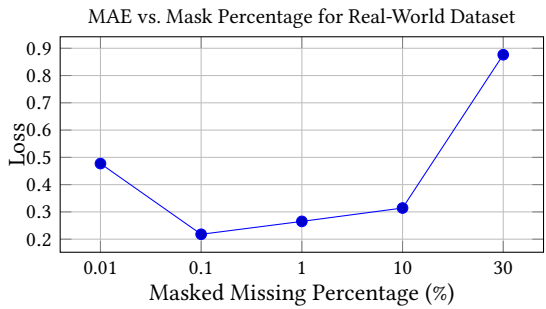
\begin{figure}[htbp]
    \centering
    \begin{tikzpicture}
        \begin{axis}[
            xlabel={Masked Missing Percentage (\%)},
            ylabel={Loss},
            xticklabels={0.01, 0.1, 1, 10, 30},
            xtick={1, 2, 3, 4, 5},
            ytick={0.2, 0.3, 0.4, 0.5, 0.6, 0.7, 0.8, 0.9},
            grid=major,
            width=0.45\textwidth,
            height=0.25\textwidth,  
            title style={yshift=-0.2cm},  
            title={MAE vs. Mask Percentage for Real-World Dataset},
            y label style={yshift=-0.2cm}, 
            x label style={yshift=0.1cm},  
            ticklabel style={font=\small}, 
            title style={font=\small}      
        ]
        \addplot coordinates {
            (1, 0.4772)
            (2, 0.2179)
            (3, 0.2652)
            (4, 0.3142)
            (5, 0.8763)
        };
        \end{axis}
    \end{tikzpicture}
    \vspace{-0.4cm}
    \caption{ An example of the U-shaped relationship between selected mask percentage and model performance with fixed level of projected missing data observed from a series of training rounds on a CA Water Dataset.}
    \label{fig:loss_vs_mask}
    \Description{}
\end{figure}

\subsection{Determining Minimal Effective Mask}
\label{sec:min_mask}
A key aspect of our training strategy involves applying a degree of additional artificial masking after an initial projection of realistic missing patterns (derived from $\incompleteTScluster[]{}$) onto complete data (from $\completeTScluster[]{}$). Through our empirical findings with this two-stage approach, we observed that imputation model performance concerning validation loss often follows a U-shaped curve (as exemplified in Figure~\ref{fig:loss_vs_mask}). Initially, moderate levels of this additional artificial masking tend to enhance model performance; we attribute this to the masking acting as a regularizer or data augmentation technique, compelling the model to learn more robust features beyond the specific projected patterns and improving generalization to diverse missingness scenarios. However, as the rate of additional artificial masking becomes excessive, performance progressively degrades.

This U-shaped performance characteristic necessitates a systematic search for a minimal effective mask ($\masksize{}^*$), representing the optimal balance between induced robustness and information preservation. A key observation across the infrastructure settings we explored is that this minimum effective mask ($\masksize{}^*$) is typically found at relatively small percentages; applying more masking beyond this optimal point generally increases computational costs for training without commensurate performance benefits. To determine $\masksize{}^*$ for a given dataset context prepared with projected patterns, Algorithm~\ref{alg:mask_search} implements a logarithmic sampling strategy that concentrates evaluations in the lower range of mask percentages, where $\masksize{}^*$ is often located, by starting from an initial $\masksize{\text{min}} = 0.01$ and evaluating mask sizes $\masksize{i} = \masksize{\text{min}}(1 + r)^i$, where $r$ controls the growth rate.

For each candidate mask size $\masksize{}$ two training scenarios are considered. First, an oracle model \( M_{\text{oracle}} \) is trained using only $\projectedTScluster[mask]{k}$ with a masking percentage $\masksize{i}$, representing the best achievable performance when training with complete sequences. Second, a model \( M_{\text{real}} \) is trained using $\projectedTScluster[proj,mask]{k}$, which incorporates both projected patterns and masking. Both models compute their losses \( L_{\text{oracle}}(m_i) \) and \( L_{\text{real}}(m_i) \) on a held-out validation set \( V_k \subset \completeTScluster{k}\) using the model's loss function \( \ell(\cdot,\cdot) \). The absolute difference between these losses defines a gap that measures how well the model trained with projected patterns matches the performance of the oracle. To account for training variability, a convergence threshold \( \omega = 2\sigma_{\text{oracle}} \) is established, where \( \sigma_{\text{oracle}} \) is the standard deviation of \( L_{\text{oracle}} \) in multiple training runs with random initializations. When the gap falls below \( \omega \), the minimum mask size is selected.

\begin{algorithm}
\caption{Logarithmic Minimum Mask Search}
\label{alg:mask_search}
\small
\DontPrintSemicolon
\KwIn{Initial mask $\masksize{min} = 0.01$, growth rate $\alpha$, threshold $\omega$, int $T$}
\KwOut{Minimum effective mask percentage $m^*$}
gaps $\gets$ [ ] \\
\For{$i \gets 0$ \KwTo $T-1$}{
   $\masksize{i} \gets \masksize{min}(1 + \alpha)^i$ \\
   Train models on $\projectedTScluster{mask}{k}$ and $\projectedTScluster[proj,mask]{k}$ with $\masksize{i}$ \\
   gap $\gets |L_{\text{oracle}} - L_{\text{real}}|$ \\
   gaps.append(gap) \\
   \lIf{gap $< \omega$}{
       $m^* \gets \masksize{i}$ \Return
   }
}
$m^* \gets \masksize{\text{argmin(gaps)}}$ \\
\Return $m^*$
\end{algorithm}

\subsection{Statistical Learning Guarantees}
\label{sec:pac_analysis}

Section \ref{sec:min_mask} introduced an algorithm for finding a minimal effective mask, but this empirical approach raises an important question: \textit{"Given that data is truly missing, how can confidence in the cluster model quality be established?"} The challenge stems from test validation: when true values are missing in production data, standard error metrics on held-out sets cannot verify model performance on the missing labels. However, by projecting patterns from $\projectedTScluster[proj]{k}$ onto complete sequences in $\completeTScluster{k}$ the model's ability to reconstruct missing values can be directly evaluated against known ground truth values. For cluster $k$, let $S_k \subseteq \completeTScluster{k}$ be the set of complete sequences that have had patterns from $\projectedTScluster{k}$ projected onto them during training. For each window $\window$ in these sequences, the reconstruction ability of the model can be evaluated through the test statistic:
\begin{equation}
T_k(\window) = \begin{cases}
1, & \text{if } |\imputationmodel{k}(\window) - y_i| \leq \tau \text{ for projected values } y_i \\
0, & \text{otherwise}
\end{cases}
\end{equation}

where $\imputationmodel{k}$ is the imputation model for cluster $k$, and $\tau$ is the reconstruction error tolerance. This tolerance represents the maximum acceptable difference between predicted and true values, typically set as a percentage of the data standard deviation (e.g. $\tau = 0.1\sigma$ means predictions within 10\% of the standard deviation) \cite{devlin2019bert}. By converting the continuous regression problem into a binary classification (success/failure) via this threshold, the test statistic enables the application of PAC learning theory \cite{valiant1984theory} to bound the probability of learning an effective imputation model:
\begin{equation}
\begin{aligned}
\alpha_k &= \frac{1}{|S_k| \cdot \window} \sum_{\TS{i}{} \in S_k} (1 - \maskproj) &\text{(proportion of projected values)} \\
\beta_k &= m^* & \text{(proportion of masked values)} \\
\gamma_k &= 1 - \alpha_k - \beta_k &\text{(proportion of observable values)}
\end{aligned}
\end{equation}

Here, \( \alpha_k \) measures the average proportion of projected values in the training set, \( \beta_k \) represents the proportion of values masked during training using the minimal effective mask \( m^* \), and \( \gamma_k \) represents the remaining observable proportion. Note that \( \alpha_k \) is scaled from window to window of data, while the mask \( m^* \) is fixed across all windows. Let \( \gamma_{\text{min}} \) be the minimum threshold of observable values required for training. The constraint \( \gamma_k \geq \gamma_{\text{min}} \) defines a feasible region for learning \cite{vapnik1971uniform}. The classification of whether a given configuration allows successful learning is as follows:
\begin{equation}
f_k(\alpha_k, \beta_k) = \begin{cases}
1 & \text{if } \mathbb{P}_{x \sim D^+_k}[T_k(x) = 1] \geq 1 - \delta \\
0 & \text{otherwise}
\end{cases}
\end{equation}

This creates a decision boundary in the \( (\alpha_k, \beta_k) \) space that separates configurations that have reliable learning from those that do not \cite{vapnik1998statistical}. The problem of finding this boundary can be viewed as learning a hypothesis from the class:
\begin{equation}
\mathcal{H}_k = \{ h_{\theta} : h_{\theta}(\alpha_k, \beta_k) = \mathbb{I}(\theta_1 \alpha_k + \theta_2 \beta_k + \theta_3 \geq 0) \text{ for } \theta \in \mathbb{R}^3 \}
\end{equation}

where \( \mathbb{I} \) is the indicator function and $\theta_3$ is a constant term. This hypothesis class is an artifact of PAC learning, corresponding to a linear classifier in the \( (\alpha_k, \beta_k) \) space with a VC dimension of 3 \cite{vapnik1971uniform, blumer1989learnability}. For any three points in this space, a hyperplane exists that realizes all \( 2^3 \) possible binary classifications, i.e., if the statistical test is satisfied. However, it is impossible to shatter four points with a hyperplane in \( \mathbb{R}^2 \), as demonstrated by the XOR configuration \cite{minsky1969perceptrons}. Applying PAC learning theory with the VC dimension of 3 yields the following bound for the minimum number of observable sequences needed for reliable learning in cluster \( k \) \cite{valiant1984theory, blumer1989learnability}:
\begin{equation}
|S_k| \gamma_k \geq \frac{1}{\epsilon} \left( 3 \log \frac{1}{\epsilon} + \log \frac{1}{\delta} \right)
\end{equation}

where \( \epsilon \) is the error tolerance for learning the decision boundary and \( \delta \) is the failure probability. When a cluster's observable size falls below this threshold, reliable learning cannot be guaranteed with probability \( 1 - \delta \) \cite{vapnik1998statistical}. These bounds establish probabilistic guarantees about the \textbf{minimum amount of observable data} needed to generalize performance of the projected values, which serve as an approximation of truly missing data.
\section{Evaluation}
\label{sec:eval}
This section evaluates DIM-SUM, which serves as a preprocessing framework applicable to any existing imputation model. Our evaluation specifically applies DIM-SUM to 5 distinct imputation-specific models (MRNN, SAITS, Nonstationary Transformer, StemGNN, and CSDI \cite{yoon2018estimating, du2023saits, liu2022non, cao2020stemgnn, tashiro2021csdi}), studying their performance across 6 datasets. These datasets span real-world and benchmark scenarios within the water, electricity, and weather infrastructure domains.

We experimentally investigate DIM-SUM's ability to enhance model accuracy by leveraging information from missing data patterns. This is examined across diverse levels and types of missingness \textbf{(Exp 1 \& 2)}. For our real-world datasets (WD1, WD2, GESL-V, GESL-C, Weather) with inherent missingness, DIM-SUM's methodology involves projecting these naturally observed patterns onto complete data segments. As a benchmark, we generate and apply synthetic Missing Not at Random (MNAR) and Missing Completely at Random (MCAR) patterns for controlled experimentation. Furthermore, we compare the performance of DIM-SUM when applied to these datasets against the strategy of adapting a large pre-trained model \textbf{Exp 3}, demonstrating DIM-SUM's potential to offer competitive results with greater efficiency.

Beyond these comparisons, we evaluate DIM-SUM against another preprocessing framework, DAGAN \cite{liu2021adaptive}, which also exploits missing data patterns prior to model training \textbf{(Exp 4)}. Lastly, we provide a comprehensive assessment of DIM-SUM’s performance across all combinations of model architectures and compared techniques, focusing on critical metrics such as computational running-time and reduction in required training data \textbf{(Exp 5 \& 6)}.

\subsection{Datasets}
\label{sec:evaluation_datasets}

In Table~\ref{tab:datasets}, we provide detailed information on the 6 datasets used in our experiments. These consist of Advanced Metering Infrastructure (AMI) data from CA cities, other infrastructure-related data, and a weather reanalysis dataset. 

\begin{table}[H]
\caption{Overview of Evaluation Datasets}
\vspace{-0.3cm}
\label{tab:datasets}
\centering
\small 
\begin{tabular}{@{}l|l@{}}
\toprule
\textbf{Infrastructure Dataset} & \textbf{Characteristics} \\
\midrule
\textbf{Water District 1 (WD1)} & 1.6B readings, 68,000 meters\\
Interval: 15 min & Duration: 2 years\\
missing: 30\% & Feature: Flow rate\\
\midrule
\textbf{Water District 2 (WD2)} & 450M readings, 18,000 meters\\
Interval: 15 min & Duration: 2 years\\
missing: 47\% & Feature: Flow rate\\
\midrule
\textbf{London Smart Meters (LCL)} & 168M readings, 360 sources\\
Interval: 10 min & Duration: 1 year\\
missing: 10-90\% & Feature: EV charging load\\
\midrule
\textbf{GESL-Current (CESL-C)} & 11M readings, 13 meters\\
Interval: 0.033s & Duration: One month\\
missing: 10-90\% & Feature: Grid current\\
\midrule
\textbf{GESL-Voltage (GESL-V)} & 11M readings, 13 meters\\
Interval: 0.033s & Duration: One month\\
missing: 10-90\% & Feature: Grid voltage\\
\midrule
\textbf{Weather} & 43M readings\\
Interval: 60 min & Duration: One year\\
missing: 10-90\% & Feature: Climate\\

\bottomrule
\end{tabular}
\end{table}

\begin{figure}[H]
    \centering
    \includegraphics[width=0.40\textwidth]{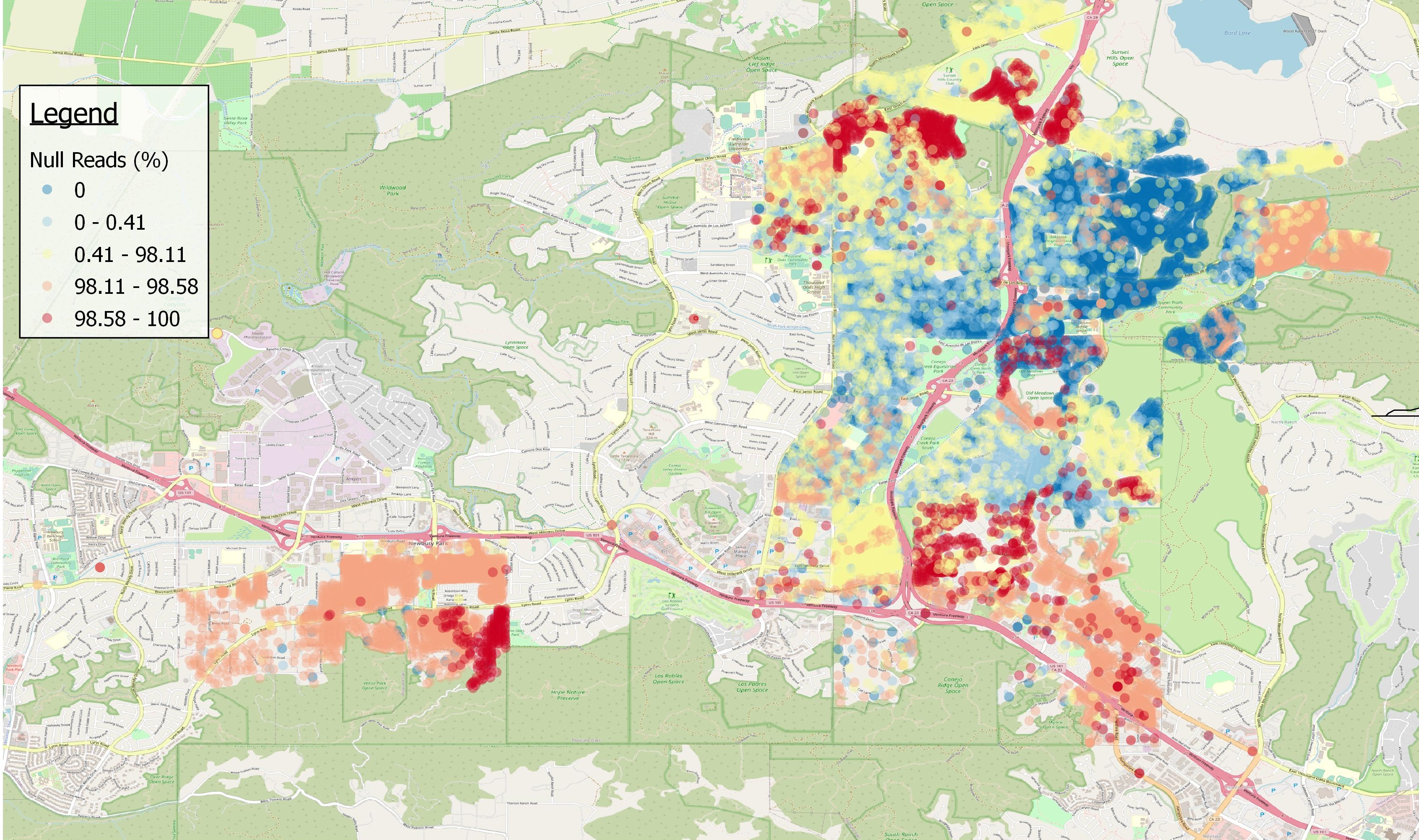}
    \caption{A Map of WD2 Displaying the Percentage of Missing Readings from Meter Sources (Red: 98\%+ missing values) }
    \label{fig:meter_quality}
    \Description{}
\end{figure}

\paragraph{\textbf{Water Consumption Datasets (WD1, WD2)}}
We utilize data from two CA water districts collected over a two-year period. These represent real-world infrastructure data with naturally occurring missing values. WD1 contains readings from 68,000 AMI meters (1.6 billion measurements), while WD2 comprises 18,000 meters with 450 million readings. Both datasets record flow values at 15-minute intervals. We observe that 30\% and 47\% of values are missing in WD1 and WD2, respectively. WD1 has been anonymized and released alongside this paper in~\cite{hildebrant_2025_wd1}. Fig.~\ref{fig:meter_quality} illustrates the variability in the amount of missing data in various geographical regions. 

\paragraph{\textbf{Electricity Usage Datasets (LCL, GESL-V, GESL-C)}} For electricity usage, we examine the London Smart Meters dataset from the Low Carbon London (LCL) project~\cite{Tindemans2023}, containing 168 million readings from 360 sources at 10-minute intervals. We also use the Grid Event Signature Library (GESL)~\cite{ORNL_gesl} with PMU measurements at 30 Hz over 3-minute intervals for current (GESL-C) and voltage magnitude (GESL-V). These datasets are typically more complete; thus, for controlled experiments, missingness is often induced.

\paragraph{\textbf{Weather Benchmark Dataset}}
To further validate our approach on benchmarks, we utilize a weather dataset, ERA5 \cite{Nguyen2023ClimateLearn}. The dataset measures hourly estimates for a large number of atmospheric, land, and oceanic climate variables over the past 80 years. We sample 43 million points from this dataset, representing a year of data. 

We process all datasets into uniform sequence lengths of 96 (corresponding to 24-hour windows at 15-minute intervals, or an 4 days for hourly intervals). For water consumption datasets with observable missing patterns, DIM-SUM directly projects these patterns onto complete data segments. For electricity and weather datasets where observations are complete, we apply MNAR and MCAR masking at specified percentages. We use a uniform 20\% holdout set for evaluation for all model and dataset combinations.

\subsection{Models and Strategies}  
\label{sec:experimental_setup}

To comprehensively evaluate DIM-SUM and understand its interaction with various imputation strategies, we first conducted a broad study across a diverse set of imputation model architectures. We examined the impact of these architectures when applied to different methodologies, including a standard direct application of each imputation model, adjusting the preprocessing to be missing pattern aware (DAGAN \cite{liu2021adaptive} \& DIM-SUM), and utilizing a pre-trained model for inference-based imputation (Chronos \cite{ansari2024chronos}).

\paragraph{\textbf{Imputation Model Architectures Studied:}}
Our initial investigation covered 8 state-of-the-art time-series imputation models, chosen to represent a range of techniques from recurrent and transformer-based to graph neural networks and diffusion models:
\begin{itemize}

\item \textbf{MRNN} (Multi-directional Recurrent Neural Networks for Time Series) \cite{yoon2018estimating}: \textbf{Recurrent architecture} that processes temporal data from multiple directions.

\item \textbf{BRITS} (Bidirectional Recurrent Imputation for Time Series) \cite{cao2018brits}: Bidirectional \textbf{recurrent neural network}, with a focus on leveraging correlations in missingness.

\item \textbf{SAITS} (Self-Attention-based Imputation for Time Series)~\cite{du2023saits}: A multi-staged \textbf{transformer architecture} with diagonally-masked self-attention blocks and feature reconstruction, designed specifically for imputation.

\item \textbf{Nonstationary Transformer}~\cite{liu2022non}: A modified \textbf{transformer} with series stationarization and de-stationary attention mechanisms that explicitly model temporal variations and non-stationarity in data distributions.

\item \textbf{Autoformer}~\cite{wu2021autoformer}: A \textbf{transformer} model featuring an auto-correlation mechanism and series decomposition blocks for enhanced time series analysis.

\item \textbf{MICN} (Multi-scale Isometric Convolutional Network) \cite{wang2023micn}: Employs multi-scale isometric \textbf{convolutions} to capture temporal patterns efficiently.

\item \textbf{StemGNN}~\cite{cao2020stemgnn}: A \textbf{Graph Neural Network (GNN)} based model that captures inter-series correlations and temporal dependencies in multivariate time series through a spectral graph perspective.

\item \textbf{CSDI}~\cite{tashiro2021csdi} (Conditional Score-based Diffusion Models for Imputation): Leverages score-based generative models conditioned on observed data to perform imputation through a \textbf{diffusion} process.

\end{itemize}

For the experimental evaluation, we selected five of these models that represent diverse and highly competitive architectural paradigms: MRNN, SAITS, Nonstationary Transformer, StemGNN, and CSDI. DIM-SUM's framework is model-agnostic, and its performance when integrated with these five models is compared against the other approaches detailed below.

\paragraph{\textbf{Strategies for Missing Pattern Integration:}}
In our experiments, we evaluate performance across different overarching strategies for handling missing data for imputation:
\begin{itemize}
\item \textbf{Direct Model Application (Standard Approach):} This approach involves \textbf{applying each imputation model architecture directly to the dataset} (which may contain inherent or induced missing values, as specified per experiment). The model is used according to its standard implementation, relying on its own internal mechanisms or recommended setup for handling and imputing missing data, without external preprocessing frameworks. This serves as a primary point of comparison to assess the added value of preprocessing frameworks.

\item \textbf{Missing Pattern Extraction Strategy (DIM-SUM):} DIM-SUM falls into this category. It first analyzes the dataset to identify and cluster data segments, then projects dominant missing data patterns onto bounded samples within each cluster for efficient and targeted model training.

\item \textbf{Pre-trained Model Strategy (Chronos):} This involves leveraging large foundation models like Chronos~\cite{ansari2024chronos}, which are \textbf{pre-trained on diverse time series data from various domains} and then applied to the specific dataset. Chronos tokenizes time series and uses a language model architecture to predict missing values based on context.

\item \textbf{Meta-Model Strategy (DAGAN):} This includes approaches like DAGAN~\cite{liu2021adaptive}, which use \textbf{generative models (e.g., GANs) to learn and replicate realistic missing data patterns}, or to directly impute values.
\end{itemize}

\subsection{Experimental Results}

We now present the detailed experimental results across the 6 datasets (WD1, WD2, LCL, GESL-V, GESL-C, and Weather) and evaluate how DIM-SUM compares to various model architectures and strategies for incorporating missing patterns.

\subsubsection{\textbf{Improvement of Existing Imputation Models using DIM-SUM}}
\label{sec:exp1_improvement}

\begin{table*}[!ht]
\setlength{\tabcolsep}{2.5pt} 
\small
\centering
\caption{MSE Comparison: Direct Model Application vs. DIM-SUM.}
\vspace{-0.3cm}
\label{tab:exp1_mcar50_model_comparison}
\begin{tabular}{@{}c|cc|cc|cc|cc|cc@{}}
\toprule
\multirow{3}{*}{\textbf{Dataset}} & \multicolumn{2}{c|}{\textbf{MRNN}} & \multicolumn{2}{c|}{\textbf{SAITS}} & \multicolumn{2}{c|}{\textbf{Nonstationary}} & \multicolumn{2}{c|}{\textbf{StemGNN}} & \multicolumn{2}{c}{\textbf{CSDI}} \\
\cmidrule{2-11}
& \textbf{Direct} & \textbf{DIM-SUM} & \textbf{Direct} & \textbf{DIM-SUM} & \textbf{Direct} & \textbf{DIM-SUM} & \textbf{Direct} & \textbf{DIM-SUM} & \textbf{Direct} & \textbf{DIM-SUM} \\
\midrule
\textbf{WD1} & 1.0887 & \textbf{0.8748} & 1.0000 & \textbf{0.9792} & 1.1491 & \textbf{1.0915} & \textbf{1.0114} & 1.2505 & 4.4240 & \textbf{3.3774} \\

\textbf{WD2} & 0.9423 & \textbf{0.8833} & \textbf{1.0027} & 1.0179 & \textbf{0.9603} & 1.0089 & 1.0798 & \textbf{1.0677} & \textbf{0.9199} & 0.9631 \\

\textbf{LCL} & 0.7190 & \textbf{0.6080} & 0.8981 & \textbf{0.6817} & 1.1910 & \textbf{1.1252} & 0.8414 & \textbf{0.8047} & 13.2458 & \textbf{6.6625} \\

\textbf{GESL-C} & 0.0079 & \textbf{0.0015} & 0.0633 & \textbf{0.0565} & 1.3024 & \textbf{1.1063} & \textbf{0.1069} & 0.1469 & 2.1340 & \textbf{0.7282} \\

\textbf{GESL-V} & 0.0007 & \textbf{0.0006} & 0.0594 & \textbf{0.0510} & 1.1122 & \textbf{1.0920} & \textbf{0.2616} & 0.9006 & 43.4276 & \textbf{26.2236} \\

\textbf{Weather} & 1.0733 & \textbf{0.8664} & 0.6229 & \textbf{0.3126} & 0.4933 & \textbf{0.3253} & 0.6135 & \textbf{0.4200} & 27.9520 & \textbf{2.6036} \\
\bottomrule
\end{tabular}
\end{table*}

This experiment evaluates the enhancement DIM-SUM brings to the five selected imputation model architectures, focusing on imputation accuracy (MSE) at fixed levels of missingness. We explicitly compare to \textbf{direct model} strategy, where the entire dataset is used for training. Table~\ref{tab:exp1_mcar50_model_comparison} presents MSE comparisons for all five models using a fixed 50\% MCAR mask and a 20\% uniform holdout dataset across each of the six datasets. 

As we saw previously, the loss achieved by a model can vary significantly from dataset to dataset. For example, the LCL dataset achieves near perfect loss using MRNN and SAITS, but skyrockets when the Non-stationary Transformer \cite{liu2022non} model. Importantly, across nearly all datasets and model configurations, DIM-SUM achieve better MSE using a significantly smaller of training data, which we discuss in \ref{sec:exp6_data_reduction}.

\subsubsection{\textbf{DIM-SUM Performance as a Function of Missing Data}}
\label{sec:exp2_missingness_function}

\begin{figure*}[!htb]
\centering
\includegraphics[width=\textwidth]{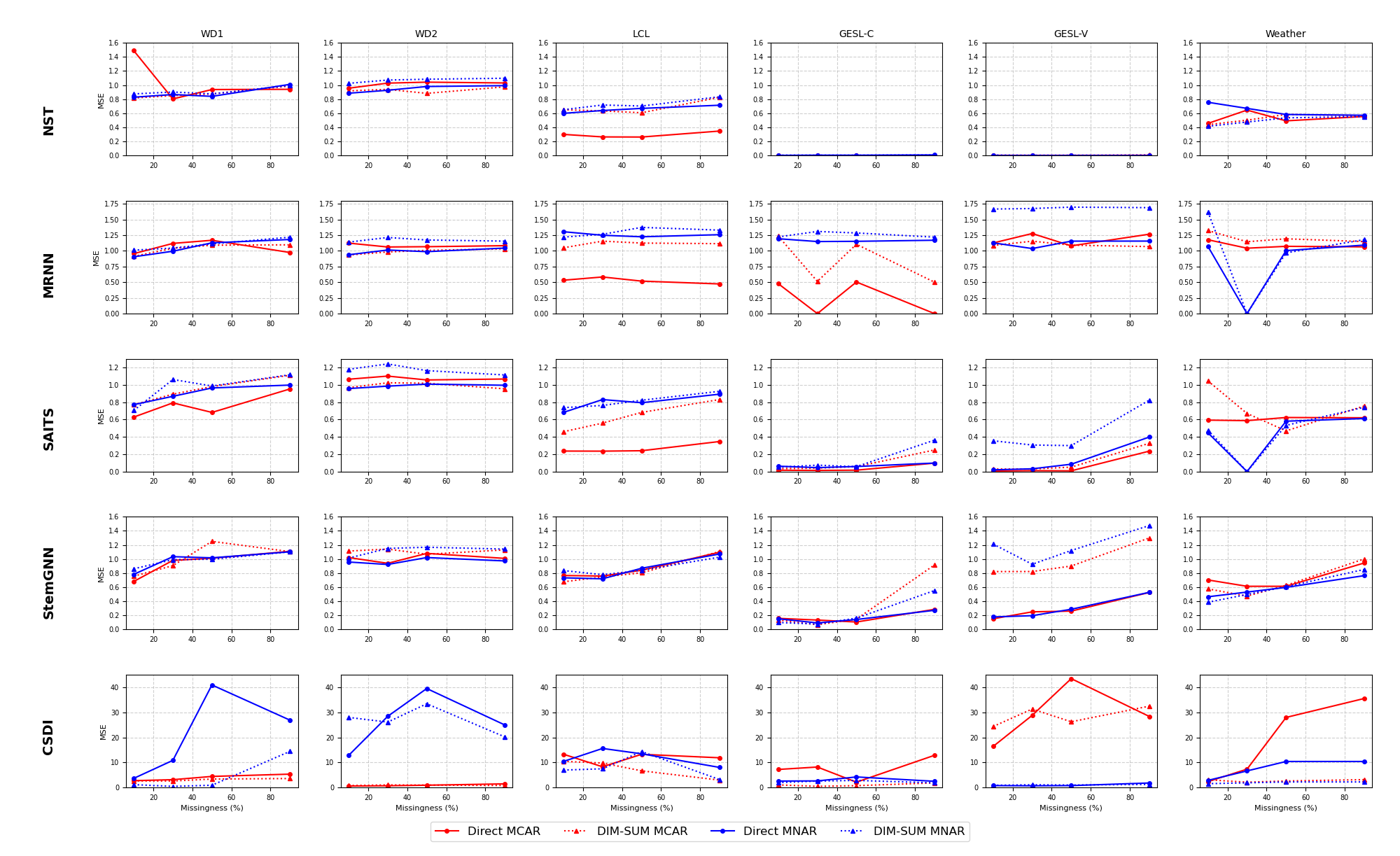}
\vspace{-0.3em}
\caption{ MSE comparison of direct models versus applying DIM-SUM across the six datasets with MCAR and MNAR}
\label{fig:mse_comparison}
\end{figure*}

DIM-SUM consistently delivers competitive imputation accuracy across the majority of datasets and throughout many missing pattern scenarios. This is demostrated through benchmarking direct application of models to various levels of missing data and missing patterns, as shown in Figure \ref{fig:mse_comparison}. Across each dataset, level of missing data, and type of missing data, DIM-SUM remains very competitive to direct applications, with the ability to train and produce results significantly faster, which we show in \ref{sec:exp5_comp_efficiency} and \ref{sec:exp6_data_reduction}. 

\subsubsection{\textbf{Comparison with a Pre-trained Model}}
\label{sec:exp3_chronos}

\begin{figure*}[!htb]
\centering
\includegraphics[width=\textwidth]{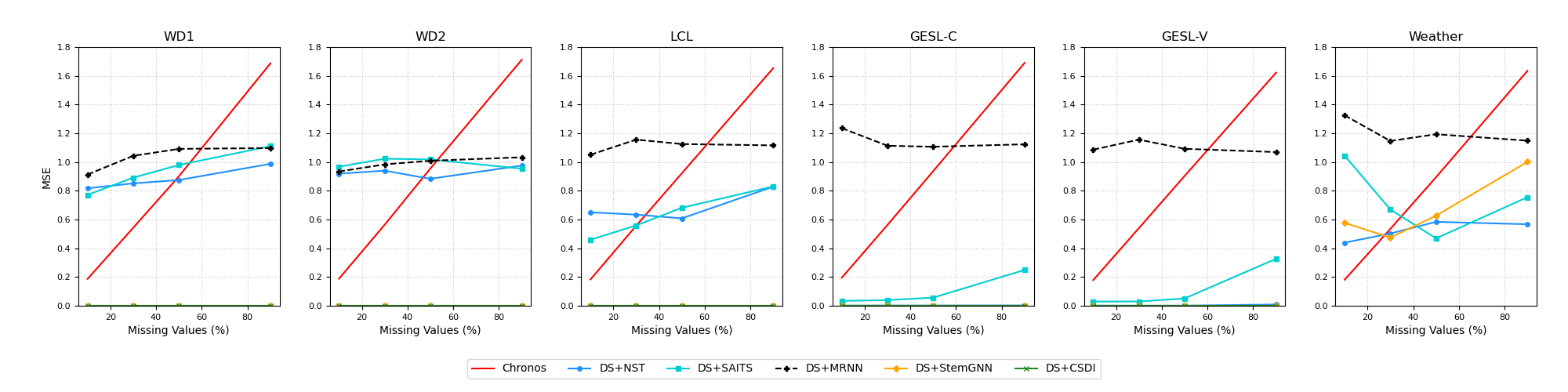}
\caption{MSE comparison of DIM-SUM applied to direct models and Chronos across the six datasets}
\label{fig:mse_comparison}
\end{figure*}

We evaluate DIM-SUM against Chronos~\cite{ansari2024chronos}, a prominent pre-trained foundation model for time series. Figure~\ref{fig:mse_comparison} illustrates comparative MSE on the 20\% holdout set for our 6 datasets. Chronos generally shows strong performance at lower missing percentages (e.g., 10-30\%), where ample context is available for its predictions. However, as missingness increases, DIM-SUM tends to exhibit more robust or superior accuracy across the datasets. For example, on WD1 at 90\% missing data, DIM-SUM achieves an MSE of 0.9874, whereas Chronos's MSE is visibly higher (1.6863). This is attributed to Chronos's sequential imputation, which can lead to error propagation when context is sparse. The significant computational cost of Chronos, especially its inference time, is discussed in Section \ref{sec:exp5_comp_efficiency}.

\subsubsection{\textbf{Comparison with a Meta-Model Strategy for missing pattern generation}}
\label{sec:exp4_dagan}

\begin{table*}[!h]
\centering
\caption{MSE Comparison: DAGAN vs. DIM-SUM (D / DS) across Models, Datasets, and Missingness Levels (10\% \& 90\%).}
\vspace{-0.3cm}
\label{tab:exp1_mcar_dagan_vs_dimsum_5models_comparison_10_90_consolidated_v2}
\resizebox{\textwidth}{!}{%
\begin{tabular}{@{}c|cc|cc|cc|cc|cc@{}}
\toprule
\multirow{3}{*}{\textbf{Dataset}} & \multicolumn{2}{c|}{\textbf{MRNN}} & \multicolumn{2}{c|}{\textbf{SAITS}} & \multicolumn{2}{c|}{\textbf{Nonstationary}} & \multicolumn{2}{c|}{\textbf{StemGNN}} & \multicolumn{2}{c}{\textbf{CSDI}} \\
\cmidrule{2-11}
& \textbf{@10\%} & \textbf{@90\%} 
& \textbf{@10\%} & \textbf{@90\%} 
& \textbf{@10\%} & \textbf{@90\%} 
& \textbf{@10\%} & \textbf{@90\%} 
& \textbf{@10\%} & \textbf{@90\%} \\
& \small(D / DS) & \small(D / DS) 
& \small(D / DS) & \small(D / DS) 
& \small(D / DS) & \small(D / DS) 
& \small(D / DS) & \small(D / DS) 
& \small(D / DS) & \small(D / DS) \\
\midrule
\textbf{WD1}     & \textbf{0.4983} / 0.9131 & 13.7267 / \textbf{1.0972} & 6.1859 / \textbf{0.7711} & 11.1561 / \textbf{1.1104} & 4.1538 / \textbf{0.8171} & 13.3162 / \textbf{0.9874} & 6.8437 / 0.7571& 18.9274 / 1.1046& 24.3528	 / 2.6256& 38.4582 / 3.5908\\
\textbf{WD2}     & 1.5169 / \textbf{0.9340} & 5.7200 / \textbf{1.0330}  & 1.8119 / \textbf{0.9673} & 5.8485 / \textbf{0.9536}  & 1.3394 / \textbf{0.9190} & 5.7613 / \textbf{0.9754}  & 1.6794 / \textbf{1.1130} & 5.8613 / \textbf{1.1297} & 5.7641 / 0.7817& 8.2235 / 0.7962\\
\textbf{LCL}     & 1.3765 / \textbf{1.0504} & 1.3756 / \textbf{1.1158}  & 0.7852 / \textbf{0.4592} & 1.0902 / \textbf{0.8304}  & 0.5174 / \textbf{0.4592} & 0.9865 / \textbf{0.8304}  & \textbf{0.1020} / 0.6773 & \textbf{0.5019} / 1.1064 & \textbf{0.6576} / 10.4309 & \textbf{0.9625} / 2.9519 \\
\textbf{GESL-V}  & 1.1524 / \textbf{1.0862} & 2.5685 / \textbf{1.0692}  & 0.0314 / \textbf{0.0293} & 3.1598 / \textbf{0.3266}  & 0.0229 / \textbf{0.0010} & 0.0315 / \textbf{0.0086}  & \textbf{0.0458} / 0.8221 & 3.2871 / \textbf{1.2994} & \textbf{0.0453} / 24.3325 & \textbf{0.6210} / 32.4632 \\
\textbf{GESL-C}  & 1.2613 / \textbf{1.2358} & 6.4517 / \textbf{1.1706}  & 0.0343 / \textbf{0.0335} & 7.9391 / \textbf{0.2492}  & 0.1231 / \textbf{0.0013} & 0.3673 / \textbf{0.0022}  & 0.1640 / \textbf{0.1354} & 8.0370 / \textbf{0.9166} & \textbf{0.0931} / 1.0443 & \textbf{0.5011} / 1.8508 \\
\textbf{Weather} & \textbf{1.0396} / 1.3247 & 11.5739 / \textbf{1.1486} & \textbf{0.1904} / 1.0443 & 13.3771 / \textbf{0.7553} & \textbf{0.3192} / 0.4389 & 4.0234 / \textbf{0.5671} & \textbf{0.2401} / 0.5769 & 13.4723 / \textbf{1.0015} & \textbf{0.1904} / 347.7666 & \textbf{0.5850} / 43.5254 \\
\bottomrule
\end{tabular}%
} 
\end{table*}

\begin{table*}[!ht]
\setlength{\tabcolsep}{2.5pt}
\small
\centering
\caption{End-to-End (E2E) Computation Time (ms) Comparison}
\vspace{-0.3cm}
\label{tab:e2e_eff_model_breakdown_increase}
\resizebox{\textwidth}{!}{%
\begin{tabular}{@{}l|l|ccc|ccc|ccc|ccc|ccc@{}}
\toprule
\multirow{2}{*}{\textbf{Dataset}} & \multirow{2}{*}{\textbf{Technique}} & \multicolumn{3}{c|}{\textbf{NST}} & \multicolumn{3}{c|}{\textbf{SAITS}} & \multicolumn{3}{c|}{\textbf{MRNN}} & \multicolumn{3}{c|}{\textbf{StemGNN}} & \multicolumn{3}{c}{\textbf{CSDI}} \\
\cmidrule{3-17}
& & \textbf{@10\%} & \textbf{@90\%} & \textbf{Incr.} & \textbf{@10\%} & \textbf{@90\%} & \textbf{Incr.} & \textbf{@10\%} & \textbf{@90\%} & \textbf{Incr.} & \textbf{@10\%} & \textbf{@90\%} & \textbf{Incr.} & \textbf{@10\%} & \textbf{@90\%} & \textbf{Incr.} \\
\midrule

\multirow{4}{*}{WD1}
& DIM-SUM & \textbf{182.0} & \textbf{194.1} & +12.1 & 292.1 & \textbf{311.4} & +19.3 & \textbf{325.9} & \textbf{311.6} & -14.3 & \textbf{366.0} & \textbf{348.6} & -17.4 & \textbf{967.9} & \textbf{963.1} & -4.8 \\
& Baseline (Direct) & 1169.4 & 1269.3 & +99.9 & 1251.9 & 1332.6 & +80.7 & 1147.1 & 1233.3 & +86.2 & 381.8 & 3343.7 & +2961.9 & 993.7 & 982.7 & -11.0 \\
& DAGAN & $5.69 \times 10^6$ & $6.13 \times 10^6$ & $+4.43 \times 10^5$ & $5.69 \times 10^6$ & $5.71 \times 10^6$ & $+1.93 \times 10^4$ & $7.47 \times 10^6$ & $7.41 \times 10^6$ & $-6.64 \times 10^4$ & $6.11 \times 10^6$ & $6.34 \times 10^6$ & $+2.30 \times 10^5$ & $6.25 \times 10^6$ & $6.31 \times 10^6$ & $+6.00 \times 10^4$ \\
& Chronos & $1.08 \times 10^5$ & $8.64 \times 10^5$ & $+7.56 \times 10^5$ & $1.08 \times 10^5$ & $8.64 \times 10^5$ & $+7.56 \times 10^5$ & $1.08 \times 10^5$ & $8.64 \times 10^5$ & $+7.56 \times 10^5$ & $1.08 \times 10^5$ & $8.64 \times 10^5$ & $+7.56 \times 10^5$ & $1.08 \times 10^5$ & $8.64 \times 10^5$ & $+7.56 \times 10^5$ \\
\midrule

\multirow{4}{*}{WD2}
& DIM-SUM & \textbf{171.1} & \textbf{173.3} & +2.2 & \textbf{263.8} & \textbf{235.6} & -28.2 & \textbf{269.7} & \textbf{276.9} & +7.2 & \textbf{373.9} & \textbf{347.1} & -26.8 & \textbf{856.0} & \textbf{869.5} & +13.5 \\
& Baseline (Direct) & 434.1 & 433.3 & -0.8 & 460.3 & 421.1 & -39.2 & 491.7 & 465.5 & -26.2 & 335.8 & 262.2 & -73.6 & 865.1 & 863.6 & -1.5 \\
& DAGAN & $5.69 \times 10^6$ & $6.13 \times 10^6$ & $+4.43 \times 10^5$ & $5.69 \times 10^6$ & $5.71 \times 10^6$ & $+1.93 \times 10^4$ & $7.47 \times 10^6$ & $7.41 \times 10^6$ & $-6.64 \times 10^4$ & $6.11 \times 10^6$ & $6.34 \times 10^6$ & $+2.30 \times 10^5$ & $6.25 \times 10^6$ & $6.31 \times 10^6$ & $+6.00 \times 10^4$ \\
& Chronos & $1.08 \times 10^5$ & $8.64 \times 10^5$ & $+7.56 \times 10^5$ & $1.08 \times 10^5$ & $8.64 \times 10^5$ & $+7.56 \times 10^5$ & $1.08 \times 10^5$ & $8.64 \times 10^5$ & $+7.56 \times 10^5$ & $1.08 \times 10^5$ & $8.64 \times 10^5$ & $+7.56 \times 10^5$ & $1.08 \times 10^5$ & $8.64 \times 10^5$ & $+7.56 \times 10^5$ \\
\midrule

\multirow{4}{*}{LCL}
& DIM-SUM & \textbf{16.2} & \textbf{14.7} & -1.5 & \textbf{199.5} & \textbf{190.6} & -8.9 & \textbf{225.4} & \textbf{227.7} & +2.3 & \textbf{29.9} & \textbf{33.0} & +3.1 & \textbf{75.2} & \textbf{72.5} & -2.7 \\
& Baseline (Direct) & 161.5 & 161.9 & +0.4 & 136.6 & 133.5 & -3.1 & 138.6 & 150.0 & +11.4 & 0.4 & 27.6 & +27.2 & 71.1 & 72.3 & +1.2 \\
& DAGAN & $5.69 \times 10^6$ & $6.13 \times 10^6$ & $+4.43 \times 10^5$ & $5.69 \times 10^6$ & $5.71 \times 10^6$ & $+1.93 \times 10^4$ & $7.47 \times 10^6$ & $7.41 \times 10^6$ & $-6.64 \times 10^4$ & $6.11 \times 10^6$ & $6.34 \times 10^6$ & $+2.30 \times 10^5$ & $6.25 \times 10^6$ & $6.31 \times 10^6$ & $+6.00 \times 10^4$ \\
& Chronos & $1.08 \times 10^5$ & $8.64 \times 10^5$ & $+7.56 \times 10^5$ & $1.08 \times 10^5$ & $8.64 \times 10^5$ & $+7.56 \times 10^5$ & $1.08 \times 10^5$ & $8.64 \times 10^5$ & $+7.56 \times 10^5$ & $1.08 \times 10^5$ & $8.64 \times 10^5$ & $+7.56 \times 10^5$ & $1.08 \times 10^5$ & $8.64 \times 10^5$ & $+7.56 \times 10^5$ \\
\midrule

\multirow{4}{*}{GESL-C} 
& DIM-SUM & \textbf{8.5} & \textbf{8.4} & -0.1 & \textbf{9.7} & \textbf{14.5} & +4.8 & \textbf{11.8} & \textbf{14.1} & +2.3 & \textbf{21.0} & \textbf{21.2} & +0.2 & \textbf{44.0} & \textbf{42.2} & -1.8 \\
& Baseline (Direct) & 31.7 & 25.2 & -6.5 & 103.6 & 41.3 & -62.3 & 5.1 & 3.7 & -1.4 & 20.0 & 20.2 & +0.2 & 42.9 & 40.9 & -2.0 \\
& DAGAN & $5.69 \times 10^6$ & $6.13 \times 10^6$ & $+4.43 \times 10^5$ & $5.69 \times 10^6$ & $5.71 \times 10^6$ & $+1.93 \times 10^4$ & $7.47 \times 10^6$ & $7.41 \times 10^6$ & $-6.64 \times 10^4$ & $6.11 \times 10^6$ & $6.34 \times 10^6$ & $+2.30 \times 10^5$ & $6.25 \times 10^6$ & $6.31 \times 10^6$ & $+6.00 \times 10^4$ \\
& Chronos & $1.08 \times 10^5$ & $8.64 \times 10^5$ & $+7.56 \times 10^5$ & $1.08 \times 10^5$ & $8.64 \times 10^5$ & $+7.56 \times 10^5$ & $1.08 \times 10^5$ & $8.64 \times 10^5$ & $+7.56 \times 10^5$ & $1.08 \times 10^5$ & $8.64 \times 10^5$ & $+7.56 \times 10^5$ & $1.08 \times 10^5$ & $8.64 \times 10^5$ & $+7.56 \times 10^5$ \\
\midrule

\multirow{4}{*}{GESL-V} 
& DIM-SUM & \textbf{8.9} & \textbf{8.5} & -0.4 & \textbf{15.4} & \textbf{11.4} & -4.0 & \textbf{12.3} & \textbf{14.1} & +1.8 & \textbf{20.9} & \textbf{20.4} & -0.5 & \textbf{51.4} & \textbf{50.7} & -0.7 \\
& Baseline (Direct) & 27.8 & 28.1 & +0.3 & 116.5 & 36.0 & -80.5 & 6.7 & 5.7 & -1.0 & 2.3 & 20.8 & +18.5 & 50.9 & 46.8 & -4.1 \\
& DAGAN & $5.69 \times 10^6$ & $6.13 \times 10^6$ & $+4.43 \times 10^5$ & $5.69 \times 10^6$ & $5.71 \times 10^6$ & $+1.93 \times 10^4$ & $7.47 \times 10^6$ & $7.41 \times 10^6$ & $-6.64 \times 10^4$ & $6.11 \times 10^6$ & $6.34 \times 10^6$ & $+2.30 \times 10^5$ & $6.25 \times 10^6$ & $6.31 \times 10^6$ & $+6.00 \times 10^4$ \\
& Chronos & $1.08 \times 10^5$ & $8.64 \times 10^5$ & $+7.56 \times 10^5$ & $1.08 \times 10^5$ & $8.64 \times 10^5$ & $+7.56 \times 10^5$ & $1.08 \times 10^5$ & $8.64 \times 10^5$ & $+7.56 \times 10^5$ & $1.08 \times 10^5$ & $8.64 \times 10^5$ & $+7.56 \times 10^5$ & $1.08 \times 10^5$ & $8.64 \times 10^5$ & $+7.56 \times 10^5$ \\
\midrule

\multirow{4}{*}{Weather}    
& DIM-SUM & \textbf{5.1} & \textbf{6.0} & +0.9 & \textbf{10.5} & \textbf{8.5} & -2.0 & \textbf{8.0} & \textbf{9.8} & +1.8 & \textbf{12.0} & \textbf{11.3} & -0.7 & \textbf{29.4} & \textbf{28.2} & -1.2 \\
& Baseline (Direct) & 5.2 & 4.8 & -0.4 & 9.1 & 7.2 & -1.9 & 8.4 & 8.7 & +0.3 & 13.7 & 10.4 & -3.3 & 28.1 & 28.2 & +0.1 \\
& DAGAN & $5.69 \times 10^6$ & $6.13 \times 10^6$ & $+4.43 \times 10^5$ & $5.69 \times 10^6$ & $5.71 \times 10^6$ & $+1.93 \times 10^4$ & $7.47 \times 10^6$ & $7.41 \times 10^6$ & $-6.64 \times 10^4$ & $6.11 \times 10^6$ & $6.34 \times 10^6$ & $+2.30 \times 10^5$ & $6.25 \times 10^6$ & $6.31 \times 10^6$ & $+6.00 \times 10^4$ \\
& Chronos & $1.08 \times 10^5$ & $8.64 \times 10^5$ & $+7.56 \times 10^5$ & $1.08 \times 10^5$ & $8.64 \times 10^5$ & $+7.56 \times 10^5$ & $1.08 \times 10^5$ & $8.64 \times 10^5$ & $+7.56 \times 10^5$ & $1.08 \times 10^5$ & $8.64 \times 10^5$ & $+7.56 \times 10^5$ & $1.08 \times 10^5$ & $8.64 \times 10^5$ & $+7.56 \times 10^5$ \\

\bottomrule
\end{tabular}%
}
\end{table*}

Next, we compare DIM-SUM with DAGAN~\cite{liu2021adaptive}, a GAN-based framework. MSE comparisons are presented in Table~\ref{tab:exp1_mcar_dagan_vs_dimsum_5models_comparison_10_90_consolidated_v2}. DIM-SUM consistently demonstrates superior MSE compared to DAGAN, especially at higher missing percentages, across datasets. For instance, on WD1 with 90\% missingness (Non-stationary Transformer), DIM-SUM's MSE is 1.0972, markedly better than DAGAN's 13.7267. DAGAN's tendency to overfit dominant patterns contributes to this performance gap. DIM-SUM's clustering approach, in contrast, helps preserve and learn from diverse data patterns.

\subsubsection*{\textbf{Computational Efficiency}}
\label{sec:exp5_comp_efficiency}

Computational efficiency is a significant advantage of the framework. Using DIM-SUM with the following PAC bound ($\delta = 0.1, \epsilon = 0.03$), Table \ref{tab:e2e_eff_model_breakdown_increase} details the average end-to-end (E2E) computation time, directly comparing DIM-SUM with Baseline (Direct), DAGAN, and Chronos approaches at both 10\% and 90\% missingness levels. This allows for an assessment of how these techniques perform under varying data availability. DIM-SUM consistently demonstrates superior E2E efficiency. Note that full DAGAN preprocessing is impractical for large datasets. DIM-SUM's own complete preprocessing is far quicker (e.g., WD1: 45 seconds for DIM-SUM vs. 22 minutes for DAGAN on sampled data).

\subsubsection*{\textbf{Reduction of Training Data}} 
\label{sec:exp6_data_reduction}
A primary benefit of DIM-SUM is its dramatic reduction in training data requirements, as detailed in Table~\ref{tab:dataset_stats}. This table illustrates that DIM-SUM consistently operates with significantly fewer sequences compared to strategies like Direct Model Application or Chronos, which are trained using the entire available set of clean sequences. This efficiency also extends to comparisons with pattern-aware methods like DAGAN; even when DAGAN employs DIM-SUM's sampling, DIM-SUM's inclusion of majority cluster selection further refines data needs, resulting in smaller training sets. For instance, on the extensive WD1 dataset, DIM-SUM utilizes merely a fraction of the sequences needed by these alternative strategies, underscoring its superior data utilization efficiency and exceptional value, particularly in resource-constrained or large-scale deployment scenarios.

\begin{table}[!ht]
\setlength{\tabcolsep}{3pt}
\small
\centering
\caption{Reduction in training data required for DIM-SUM}
\vspace{-3mm}
\begin{tabular}{c|cccc|c}
\hline
\textbf{Dataset} & \textbf{\begin{tabular}[c]{@{}c@{}}Clean Seq.\\ (Baseline)\end{tabular}} & \textbf{Total Seq.} & \textbf{DAGAN} & \textbf{DIM-SUM} & \textbf{\begin{tabular}[c]{@{}c@{}}Reduction\end{tabular}} \\ \hline
WD1    & 4,888,091 & 16,666,667 & 240,000 & 58,025 & 189-287x \\
WD2    & 635,274   & 4,888,091  & 100,000 & 51,241 & 12.4-95x \\
LCL    & 1,746,141 & 1,746,141  & 40,000  & 7,969  & 219x     \\
GESL-V & 115,622   & 115,622    & 20,000  & 2,514  & 96x      \\
GESL-C & 115,622   & 115,622    & 20,000  & 2,514  & 96x      \\ 
Weather & 427,083   & 427,083    & 40,000  & 1,564 & 273x      \\ \hline
\end{tabular}    \label{tab:dataset_stats}
\vspace{-0.3cm}
\end{table}

\section{Conclusion}
\label{sec:conclusion}
DIM-SUM presents an innovative and scalable preprocessing framework designed to enhance the training of imputation models for time series data, particularly when faced with extensive missing values and limited complete data.  By integrating pattern clustering, adaptive masking, and statistical learning guarantees, DIM-SUM effectively navigates the challenges posed by real-world missing data patterns, moving beyond traditional artificial masking techniques.  This approach allows existing imputation models to learn from and adapt to the diverse and complex missing data scenarios frequently encountered in domains like smart utility management.

The framework's efficacy is demonstrated through comprehensive experiments across six diverse datasets, including water, electricity, and weather data, utilizing five different imputation model architectures.  These evaluations show that DIM-SUM not only achieves comparable or improved accuracy against baseline and state-of-the-art methods but does so with significantly reduced training data and faster processing times.  Furthermore, DIM-SUM provides theoretical learning guarantees, offering reliable performance even with minimal clean training data, making it a valuable tool for practical, large-scale time series applications. 

\begin{acks}
    This work is supported by NSF Grant No. 1952247. Special thanks to the California Data Collaborative and California Water Districts.
\end{acks}


\bibliographystyle{ACM-Reference-Format}
\bibliography{thebib}

\end{document}